\pdfoutput=1
\documentclass[10pt,twocolumn,letterpaper]{article}

\usepackage{iccv}
\usepackage{times}
\usepackage{epsfig}
\usepackage{graphicx}
\usepackage{amsmath}
\usepackage{amssymb}
\usepackage{booktabs}
\usepackage{threeparttable}
\usepackage{bigstrut,multirow,rotating}
\usepackage{makeidx}
\usepackage{epstopdf}

\usepackage[pagebackref=true,breaklinks=true,letterpaper=true,colorlinks,bookmarks=false]{hyperref}

\iccvfinalcopy 

\def\iccvPaperID{8918} 

\ificcvfinal\fi

\begin{document}

\title{More Separable and Easier to Segment:  A  Cluster Alignment Method for  Cross-Domain Semantic Segmentation}

\author{Shuang Wang, Dong Zhao, Yi Li, Chi Zhang, Yuwei Guo, Qi Zang, Biao Hou, Licheng Jiao \\
School of Artificial Intelligence, Xidian University, Shaanxi, China\\
{\tt\small shwang@mail.xidian.edu.cn}
}

\maketitle
\ificcvfinal\thispagestyle{empty}\fi

\begin{abstract}
Feature alignment between domains is one of the mainstream methods for Unsupervised Domain Adaptation (UDA) semantic segmentation. 
Existing feature alignment methods for semantic segmentation learn domain-invariant features by adversarial training to reduce domain discrepancy, but they have two limits: 
1) associations among pixels are not maintained,
2) the classifier trained on the source domain couldn't adapted well to the target.
In this paper, we propose a new UDA semantic segmentation approach based on domain closeness assumption to alleviate the above problems.
Specifically, a prototype clustering strategy is applied to cluster pixels with the same semantic, 
which will better maintain associations among target domain pixels during the feature alignment.
After clustering, to make the classifier more adaptive, a normalized cut loss based on the affinity graph of the target domain is utilized, 
which will make the decision boundary target-specific.
Sufficient experiments conducted on GTA5 $\rightarrow$ Cityscapes and SYNTHIA $\rightarrow$ Cityscapes proved the effectiveness of our method, which illustrated that our results  achieved the new state-of-the-art.
\end{abstract}


\section{Introduction}

\begin{figure}[h]
\includegraphics[width=0.48\textwidth, height=0.31\textwidth]{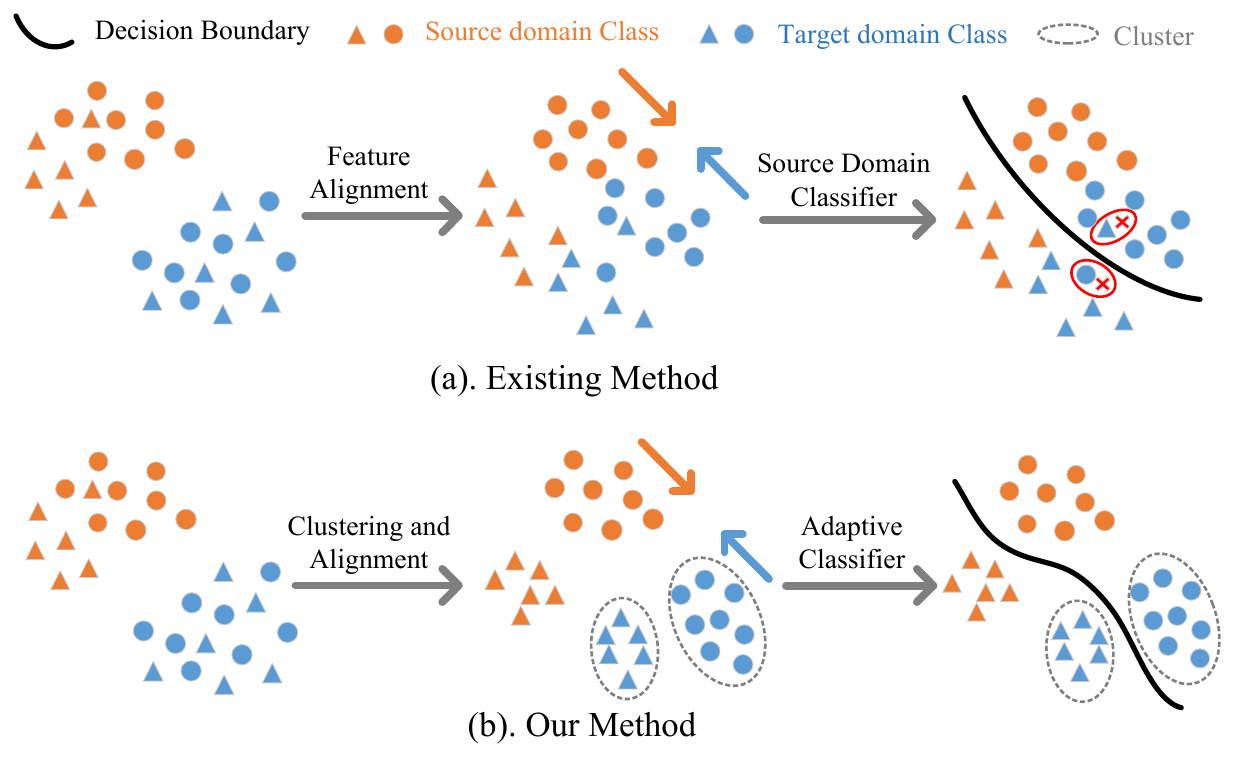}
\caption{Comparison of existing and our method.
Existing methods first align features then use the classifier trained on source domain for target domain directly. 
Our method firstly clusters the target domain features, then aligns clusters to the source, finally exploits an adaptive classifier to segmentation. (Best viewed in colour.)
}
\label{fg1}
\centering
\end{figure}

As one of the key problems in the field of computer vision, studies about semantic segmentation \cite{fcn, deeplab_v2, deeplab_v3_plus} have a remarkable progress with the development of deep neural networks. 
Semantic segmentation needs to inference a category label for each pixel in an image, and manually labeled data for training at pixel level is time-consuming, so some researchers exploit synthetic data (such as computer-rendered images) as training data, which can provide pixel-level labels conveniently \cite{gta5_dataset, syhth_dataset}. 
However, due to the difference between synthetic data and real scene data, such as perspective, scene layout, illumination, shadows and textures, etc \cite{Curriculum_da}, the trained model don't work well in the real scene.

Unsupervised domain adaptation (UDA) is proposed to reduce the domain discrepancy between labeled source domain (synthetic data) and unlabeled target domain (real data). 
For semantic segmentation, most UDA methods attempt to align at three levels: pixel level  \cite{pmlr-v80-hoffman18a, Zhang_2018_CVPR, Yang_2020_CVPR}, feature level \cite{DBLP:journals/corr/HoffmanWYD16, Chen_2017_ICCV, Hong_2018_CVPR, Sankaranarayanan_2018_CVPR, Du_SSFDA_2019_ICCV, Luo_2019_ICCV, SIM_2020_CVPR, FADA_ECCV_2020} and semantic level \cite{Tsai_2018_CVPR, Tsai_2019_ICCV}. 
The pixel-level alignment unifies the style of images in input space. 
The feature-level learns domain-invariant feature to align in latent space. 
The semantic-level aligns spatial layout in output space. 
As the most common idea, adaptation at the feature level has been continuously developed in recent years. 
Some works \cite{DBLP:journals/corr/HoffmanWYD16, Hong_2018_CVPR, Sankaranarayanan_2018_CVPR, Luo_2019_ICCV} focus on aligning global feature distributions between domains, 
but the global distribution alignment doesn't consider the class-level joint distribution and will loss the fine-grained information for each class\cite{Luo2019TakingAC, FADA_ECCV_2020}. 
Although recent studies on class-level feature distributions alignment \cite{Luo2019TakingAC, Chen_2017_ICCV, Du_SSFDA_2019_ICCV, SIM_2020_CVPR, FADA_ECCV_2020} have some improvements, \textit{two important problems are ingored in their research}. To facilitate the discussion, we denote a segmenter $ \ G = C \circ F \ $ with a combination of feature extractor $ F $ and classifier $ C $. 

1) 
Existing feature alignment UDA segmentation methods neglect associations among pixels of the same class in the target domain, 
resulting in that these pixel features cannot be mapped gathered in the feature space.
When aligning features, some scattered target domain features of different classes may be easily mixed together and aligned to the vicinity of other classes in the source domain \cite{Luo2017LabelEL, pmlr-v80-xie18c}.

2)
Existing feature alignment methods assume that the classifier $ C $ trained on source domain can also perform well on the target if the distribution between domains is well aligned,
which merely optimizes the feature extractor $F$ to perform alignment.
Due to different characteristics of each domain and the complexity of the high-dimensional feature space, it is difficult to align the distribution between domains completely \cite{Saito_2018_CVPR, Luo_2019_ICCV}.
This assumption will make the classifier work much worse in target domain than in the source.

As shown in the Fig.\ref{fg1}.(a), 
existing methods perform feature alignment to reduce the domain discrepancy. 
With ignoring intra-domain feature separability, these methods would generate ambiguous features near the decision boundary. 
In this paper, a domain closeness assumption originated from the research in \cite{Shi_2012_Information_Theoretical, Shu2018ADA} is suggested to alleviate mentioned problems: \textit{data in both the source and the target domains are tightly clustered and clusters corresponds class boundaries. For the same class, the clusters from the two domains are geometrically close to each other}. 
Under this assumption, the alignment strategy is to reduce distance among feature clusters with the same semantic class between domains.
In this way, associations among target domain pixels can be maintained better during the alignment.
This assumption also implicitly requires the decision boundary should not go through the high-density regions, which calls for an adaptive variation for the classifier trained on source domain.
Our motivation is shown in Fig.\ref{fg1}.(b). 
Our method can maintain association of target domain features better during alignment and can make the decision boundary target-specific.

Specifically, in the source domain, rich pixel-level annotations enable the segmenter to capture the semantic associations among pixels during training, which will make the features structural distributed. 
This structural distribution is defined as s-cluster.
In the target domain, lack of labels makes it difficult to do the same thing. 
In this paper, we propose a pixel-level prototype clustering strategy to help target domain features form structural distribution, which is defined as t-cluster.
The prototype clustering assumes that a prototype exists in each class and other features tightly surround their corresponding prototypes.
As the cluster center of class, prototype is selected from features with the same pseudo-label.
And other features are clustered around the prototype by reducing the distance to the closest prototype.
After clustering, to align the feature distribution of the source and target domain, the geometric distance needs to be reduced between the t-cluster and corresponding s-cluster, which is estimated by calculating the distance between the first-order statistics of each cluster distribution.

Finally, with the clustered structure of both domains, 
the classifier should be adaptive to make the decision boundary away from the empirical data. 
A natural idea in \cite{JMLR:v7:belkin06a, graph_ssl} is used in this paper, which suggests that the corresponding labels of two samples should be the same if they are close in the high-density regions.
Thus, a feature affinity graph of the target domain is introduced to present associations among features.
Based on this graph, a normalized cut loss is applied, 
which will enable features with high affinity to be predicted as the same label by the classifier.

To summarize, our contributions are three-folds:

\begin{itemize}
\item
A novel domain adaptive semantic segmentation method based on domain closeness assumption is proposed, which can align features and maintain the associations among pixels to enhance the discriminativeness of the target domain features.
\item
A normalized cut loss is utilized to make the classifier adaptive, which can learn a target-specific decision boundary.

\item
Sufficient experiments conducted on existing benchmark tasks proved the effectiveness of our method, which illustrated that our results achieved the new state-of-the-art.
\end{itemize}

\section{Related Work}

\subsection{Domain Adaptive Semantic Segmentation}
The traditional unsupervised domain adaptation method  \cite{pmlr-v37-long15, Long2016UnsupervisedDA, pmlr-v37-ganin15, Tzeng_2017_CVPR} narrows the domain gap by reducing a certain distance metric between the source and the target domain or adversarial training. As pointed out by \cite{Curriculum_da}, this kind of method may not be suitable for semantic segmentation tasks. 
Therefore, scholars designed new unsupervised domain adaptation methods tailored to semantic segmentation.  
In \cite{pmlr-v80-hoffman18a, Zhang_2018_CVPR, Yang_2020_CVPR, Kim_2020_CVPR, Li_2019_CVPR}, they try to change visual style of the two domains, and directly reduce the domain discrepancy in the original data space. 
On the other hand, Hoffman $et$ $al$. \cite{DBLP:journals/corr/HoffmanWYD16} proposes to align the global distribution of the source and target domains at the feature space, but the improvement is relatively limited. 
Inspired by this idea, many works try to improve the feature alignment method in semantic segmentation. 
One improvement direction is to alleviate the problem of difficult alignment in high-dimensional space.
In \cite{Hong_2018_CVPR, Sankaranarayanan_2018_CVPR}, they proposed to align the two domains in the transformed low-dimensional space.
Tsai $et$ $al$. \cite{Tsai_2018_CVPR, Tsai_2019_ICCV} directly treats the output of the classifier as the transformed space, expecting to align the class layout of the two domains at the output level. Another improvement direction for feature alignment is fine-grained alignment. 
Chen and Du $et$ $al$. \cite{Chen_2017_ICCV, Du_SSFDA_2019_ICCV} design multiple class discriminators to align the class feature distribution.  
Wang $et$ $al$. \cite{SIM_2020_CVPR} considers the foreground and background categories differently, and then performs feature matching with the source domain features. 
The above-mentioned work focus on inter-domain alignment, and there are also some methods \cite{Zou_2019_ICCV, Curriculum_da, 10.1007/978-3-030-58568-6_26, Lian_2019_ICCV} that use self-training strategy for intra-domain adaptation. 

\subsection{Clustering for unsupervised domain adaptation}
Domain closeness hypothesis and clustering hypothesis are two common hypothesis, which are applied to clustering-based UDA methods.
Shi $et$ $al$. \cite{Shi_2012_Information_Theoretical} first proposes the former, which could learn discriminative cluster and exploit this structure to construct the classifier. 
Inspired by this work, Deng and Pan $et$ $al$. \cite{Deng_2019_ICCV, Pan_2019_CVPR} apply sample-level clustering strategy to align target feature efficiently.
Tang $et$ $al$. \cite{Tang_2020_CVPR} introduces a unified deep clustering framework to uncover the intrinsic discrimination among target data.
These works \cite{Shi_2012_Information_Theoretical, Deng_2019_ICCV, Pan_2019_CVPR, Tang_2020_CVPR}reveal that maintaining the structure of the target domain data is important for UDA,
but they are designed for sample-level classification tasks and are unsuitable for pixel-level semantic segmentation tasks.
Others based on clustering hypothesis \cite{Clustering_assumption} have been applied to semantic segmentation tasks.
Vu and Chen $et$ $al$. \cite{ADVENT_2019_CVPR, MaxSquare_2019_ICCV} optimize the entropy or its variant to adjust the decision boundary to cross low-density regions.
Saito $et$ $al$. \cite{Saito_2018_CVPR} aligns the distribution across domain by exploring task-specific boundaries.
Although these methods \cite{ADVENT_2019_CVPR, MaxSquare_2019_ICCV, Saito_2018_CVPR} have some improvements in semantic segmentation, they ignore the structure of the target domain data, which lead to difficulties in separating features.
We propose a new UDA semantic segmentation method based on closeness hypothesis with advantages of clustering hypothesis, which clusters the target domain features to make them separable.

\subsection{Normalized cut}
The normalized cut is proposed by \cite{NC_TPAMI_2000}, which aims at solving the image segmentation. 
Following this, 
Tang $et$ $al$. \cite{Tang_2018_ECCV, Tang_2018_CVPR} applies the normalized cut loss to CNN architecture for weakly supervised semantic segmentation, 
in which similar pixels could be classified into the same class.
These works also proved that the normalized cut loss had stable gradient and supported back propagation.
Due to the excellent properties of the normalized cut loss, we apply it for unsupervised semantic segmentation, in which features with high affinity could be classified into the same class.

\section{Method}
In this section, we describe our framework for unsupervised domain adaptive semantic segmentation, including clustering pixels with the same semantic, aligning clusters between domains and adjusting the classifier.

\subsection{Algorithm Overview}
\begin{figure*}[htp]
    \begin{center}
    \centering 
    \includegraphics[width=1\textwidth, height=0.5\textwidth]{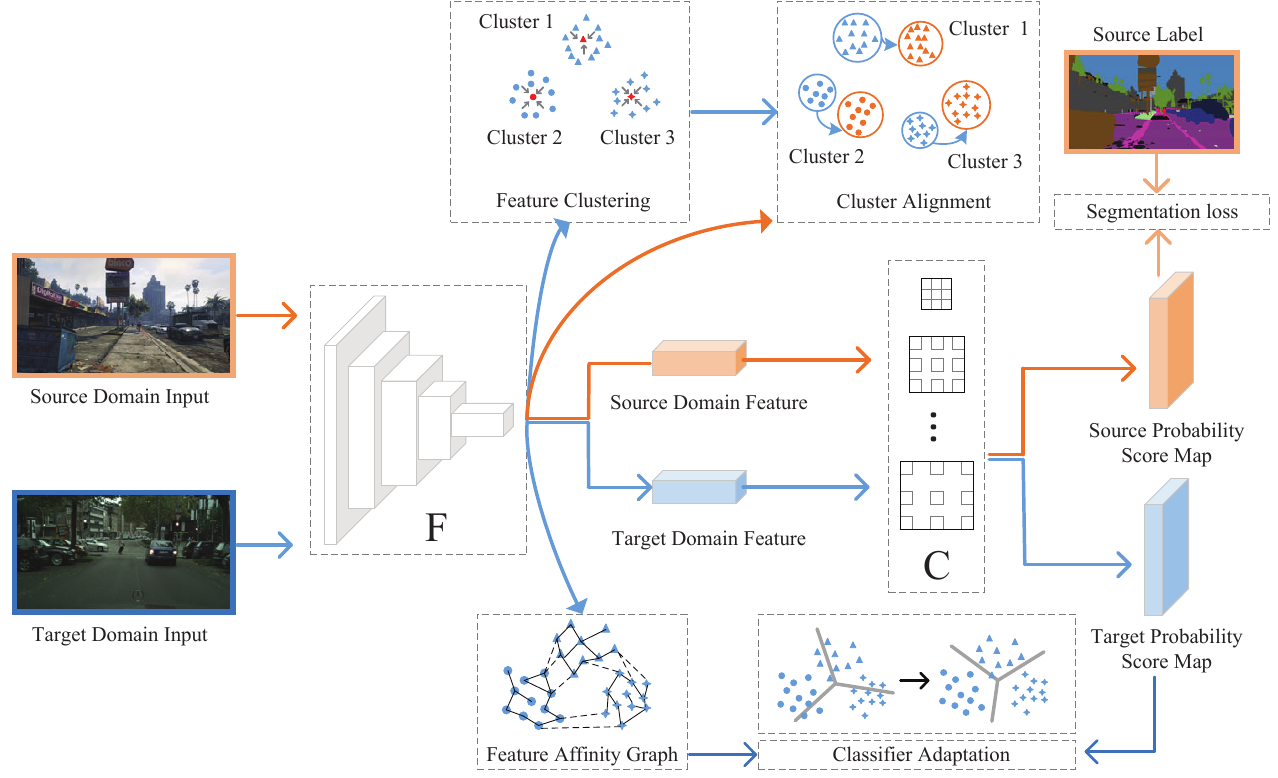} 
    \end{center}
    \caption{Overview of our proposed framework.
	    Images from both domains are fed to the feature extractor $ F $ and the classifier $ C $ to get a feature map and a probability score map.
	    For source domain (Orange), a segmentation loss is calculated between probability score map and ground truth.
	    For target domain (Blue), on the one hand, features with same semantic are clustered. After that, these clusters are aligned to the corresponding source domain cluster. 
	    On the other hand, a  feature affinity graph is built and then a normalized cut loss based on this graph is applied to the probability score map for adjusting the classifier. (Best viewed in colour.)} 
    \label{fig2}
\end{figure*}

The detail of our framework is illustrated in Fig.\ref{fig2}.
Given a set of source domain samples $ \ X_s = \{{x_{s}^{i}}\}_{i=1}^{N} \ $ with pixel-level labels  $ \ Y_s = \{{y_{s}^{i}}\}_{i=1}^{N} \ $ and  unlabeled target domain samples $ \ X_t = \{{x_{t}^{i}}\}_{i=1}^{M} \ $,
we aim at learning a segmentation model $ G $ that can work on both domains well.
It is worth noting that the target class set is the same as the source.
Generally, the segmenter $ \ G = C \circ F \ $ consists of a feature extractor $ F $:  $ X \rightarrow Z $ and a classifier $ C $: $ Z \rightarrow L $, where $ Z $ is the feature space and $ L $ is the label space. 

For the source domain, an image $ x_s  $ sampled from $ X_s  $ is sent to the segmenter $ G = C \circ F $ to get a feature map  $ f_s = F(x_s)  $ and a pixel-level probability score map $p_s = C(f_s)$.
Then a segmentation loss $ L_{seg} $ between $ p_s $ and $ y_s \in Y_s $ is used to capture the  semantic associations among source features, which makes the features belonging to the same class clustered on the feature map $ f_s $.

For the target domain, 
an image $ x_t $ sampled from $ X_t $ is also send to the segmenter $ G$ to get a feature map  $f_t$ and a pixel-level probability score map $p_t$.
According to the domain closeness hypothesis, the feature map $ f_t$ is also expected to remain structured.
Then a prototype clustering method with loss $ L_c $ is applied to target feature map $ f_t $ to maintain associations among features.
After clustering, to align target cluster to the corresponding source cluster, a contrastive alignment loss $ L_a $ is used for narrowing the domain gap.
With the aligned cluster structure, a normalized cut loss $ L_n $ is applied to make the classifier more adaptive for target domain.
Finally, the overall optimization term $ L_{full} $ is sum of all above losses. And we train the segmenter $G$ by optimizing the following target function:
\begin{equation}
\begin{aligned}
\min_{F,C}L_{full} = \min_{F}(L_{seg} + \lambda_{c} L_{c} + \lambda_{a} L_{a} + \lambda_{n} L_{n}) \\
+ \min_{C}(L_{seg} + \lambda_{n} L_{n})
 \end{aligned}
\end{equation}
where $L_{seg}$ is pixel-wise cross-entropy loss. $L_{c}$ , $ L_a $ and $ L_n $ correspond to  Section \ref{tfc},  \ref{ca} and  \ref{dbr} respectively.

\subsection{Target domain features clustering} \label{tfc}
In our method, the prototype clustering is used to maintain the association of target domain features, which assumes that a prototype exists in each class and other features tightly surround  the corresponding prototype.
For each class, the most representative feature is selected as the prototype from features with the same pseudo-label \cite{Saito_2019_ICCV}.
With given a feature map $f_{t}\in \mathbb{R}^{h \times w \times c}$ and    a probability score map $p_t \in \mathbb{R}^{h \times w \times K}$,
pseudo-label map $\hat{y_{t}} \in \mathbb{R}^{h \times w} $ is first obtained by selecting the class with the highest probability in $p_t$.
$ K $ stands for the number of classes.
$h,w,c$ stand for the width, height and channel numbers of a feature map.
And then an operation $ \sigma(f_t,  \ \hat{y_t}, k) $ is defined to select features $\{{b_{m}^{k}}\}_{m=1}^{M} $ with pseudo-label $k$.
$ M $ stands for the number of selected features.
Finally, we use the following formula to calculate the prototype $ \sigma^{k} $.
\begin{equation}
\sigma^{k} = b_{\alpha},
\end{equation}
\begin{equation}
\alpha = \arg\max_{m} \sum_{m^{\prime}=1}^{M} S(b_{m}^{k}, b_{m^{\prime}}^{k}),  
m=1,2...M,
\end{equation}
where the $S(\cdot,\cdot)$ stands for cosine similarity, which can be replaced with other similarity measures according to specific tasks.

After selecting prototypes $ \{\sigma_{k}\}_{k=1}^K $, for $i$-th ($i=1,2...w\times h$) feature $ f_{t}^{i} $ on the feature map $ f_t $, the following formula is utilized to define the conditional probability $p_{ik}$ of making $ \sigma_{k} $ be the nearest prototype of $ f_{t}^{i} $.
\begin{equation}
p_{ik} = p(y=k | f_{t}^{i})= \frac {e^{S(f_{t}^{i},  \sigma^{k})}} {\sum_{k^{\prime}=1}^{K} e^{S(f_{t}^{i},  \sigma^{k^{\prime}}) }},
\end{equation}
Then prototype clustering loss $L_c$ is used to make features get together around the corresponding prototype.
\begin{equation}
L_c = - \frac{1}{w \times h} \sum_{i=1}^{w \times h} \log p(y=k | f_{t}^{i}), 
\end{equation}

\subsection{Cluster Alignment} \label{ca}
After intra-domain feature being clustered, we need align them by classes between domains to make it easy to classify.
An approach \cite{Salimans2016ImprovedTF, Deng_2019_ICCV} is optimizing the first-order statistics between source domain cluster and corresponding target domain cluster.
However, rigorously restricting this distance may make features less  discriminative and degrade the model \cite{Xu_2019_ICCV}.
Therefore, 
the loss originated from \cite{He_MOCO_2020_CVPR} is exploited to optimize cluster distance contrastively, 
which will work better than the methods in \cite{Salimans2016ImprovedTF, Deng_2019_ICCV}.
The contrastive cluster alignment loss  $L_a$ is exploited in Eq.(\ref{eq6}), which will make the distance between target cluster and the corresponding source cluster be closer than the others.
\begin{equation}
L_{a} = -\frac{1}{K} \sum_{k=1}^{K}  \log\frac{ e^{-D(U_{t}^{k},U_{s}^{k})}} { \sum_{j=1}^{K}  e^ {-D(U_{t}^{k},U_{s}^{j})} }, \label{eq6}
\end{equation}
$ D(\cdot,\cdot) $ stands for euclidean distance, which can be replaced with other distance measures. $U_{(s,t)}^{k}$ is the normalized first-order statistics of the  $k$-th feature cluster. Here the first-order statistics are normalized by $l2$ norm to narrow the scale gap, which will facilitate the convergence of the contrast loss. 
\begin{equation}
U_{(s,t)}^{k} = \frac{u_{(s,t)}^{k}}{\|u_{(s,t)}^{k}\|_{2}} , 
\end{equation}

\begin{equation}
u_{(s,t)}^{k} =  \frac{1}{M} \sum_{i=1}^{M} {b_i^{k}}_{(s,t)}.
\end{equation}


\subsection{Classifier Adaptation}  \label{dbr}
As mentioned before, the classifier $C$ trained on source domain needs to be adjusted for adapting to target domain.
An idea in \cite{JMLR:v7:belkin06a, graph_ssl} is utilized, which suggests that features with high affinity are expected to be given the same label.
Since the decision boundary should not separate them, 
it conforms to our assumption that the decision boundary should not go through high-density regions. 
Thus, we introduce feature affinity graph to present the association among features of each feature map $f_t$.
The affinity matrix $A \in \mathbb{R}^{hw \times hw} $ of the graph is defined as:
\begin{equation}
A_{ij} = \frac {e^{S(f_t^i, f_t^j)}} {\sum_{j^{\prime}}e^{S(f_t^i, f_t^{j^{\prime}})}}.
\end{equation} 

Then the spectral clustering \cite{2001On} can be utilized to assign cluster label to each node in the affinity graph. 
These graph node labels better maintain the structure of target domain features, and can be used to adjust the classifier.
However, this operation requires iteration and is difficult for applying to neural networks.
Tang $et$ $al$. \cite{Tang_2018_CVPR} proposes a normalized cut loss, 
which has a similar function with spectral clustering and can be directly applied to neural networks. 
Therefore, in our method, the normalized cut loss $L_n$ is applied 
and is defined in Eq.(\ref{eq10}).
\begin{equation}
L_n = \sum_k \frac {cut(\Omega, \Omega/\Omega_k)} {assoc(\Omega_k, \Omega)}
=\sum_{k}^{K}  \frac {p_{t}^{k}A(1 - p_{t}^{k})} {d^{'} p_{t}^{k}} , \label{eq10}
\end{equation}
where $d$ is the degree vector of matrix $A$ and $ p_{t}^{k}$ is the probability score vector of the $k$-th class.

Minimizing $L_n$ is to find a set of cuts on the affinity graph, which can make the disassociation among partitioned K-subgraphs minimized and the association within subgraphs maximized \cite{NC_TPAMI_2000}.
In other words, this set of cuts works as a decision boundary of crossing the low-density regions.
And $p_{t}$ provides an initial soft label of the graph, 
which makes the graph nodes label the same as the segmentation label.
Therefore, optimizing $L_n$ can directly adjust the semantic segmentation results and can pass the gradient back to the classifier.


\begin{table*}[htbp]
  \centering
  \caption{Experimental results for GTA5 $\rightarrow$ Cityscapes. For fair comparison, we classify the existing methods into three types: input image translation, feature alignment and self-training, which are defined as "I", "F" and "S" respectively. Our method is not only compared with feature alignment methods, but also with multi-strategy methods.}
  \resizebox{\textwidth}{!}{
    \begin{tabular}{c|c|ccccccccccccccccccc|c}
    \toprule
    \multicolumn{22}{c}{GTA5-Cityscapes} \\
    \midrule
          & M     & Road  & SW    & Build & Wall  & Fence & Pole  & TL    & TS    & Veg   & Trrn  & Sky   & PR    & Rider & Car   & Truck & Bus   & Train & Motor & Bike  & mIoU \\
    \midrule
    \multicolumn{1}{l|}{Source Only} &       & 48.8  & 22.4  & 65.5  & 24.5  & 15.2  & 26.9  & 30.2  & 14.3  & 82.2  & 24.7  & 77.1  & 56.4  & 34.5  & 55.1  & 11.9  & 23.9  & 0.1   & 26.5  & 21.3  & 34.8  \\
    \midrule
    SIBAN\cite{Luo_2019_ICCV} & F     & 88.5  & 35.4  & 79.5  & 26.3  & 24.3  & 28.5  & 32.5  & 18.3  & 81.2  & 40.0  & 76.5  & 58.1  & 25.8  & 82.6  & 30.3  & 34.4  & 3.4   & 21.6  & 21.5  & 42.6  \\
    CLAN\cite{Luo2019TakingAC}  & F     & 87.0  & 27.1  & 79.6  & 27.3  & 23.3  & 28.3  & 35.5  & 24.2  & 83.6  & 27.4  & 74.2  & 58.6  & 28.0  & 76.2  & 33.1  & 36.7  & 6.7   & 31.9  & 31.4  & 43.2  \\
    SSF-DAN\cite{Du_SSFDA_2019_ICCV} & F     & 90.3  & 38.9  & 81.7  & 24.8  & 22.9  & 30.5  & 37.0  & 21.2  & 84.8  & 38.8  & 76.9  & 58.8  & 30.7  & 85.7  & 30.6  & 38.1  & 5.9   & 28.3  & 36.9  & 45.4  \\
    FADA\cite{FADA_ECCV_2020}  & F     & 90.4  & 40.3  & 83.6  & 37.6  & 17.9  & 29.8  & 33.5  & 20.6  & 83.0  & 37.9  & 84.5  & 58.9  & 35.3  & 85.0  & 38.4  & 44.7  & 20.7  & 25.2  & 23.0  & 46.9  \\
    CRST\cite{Zou2019ConfidenceRS}  & S     & 91.0  & 55.4  & 80.0  & 33.7  & 21.4  & 37.3  & 32.9  & 24.5  & 85.0  & 34.1  & 80.8  & 57.7  & 24.6  & 84.1  & 27.8  & 30.1  & 26.9  & 26.0  & 42.3  & 47.1  \\
    MaxSquare\cite{ MaxSquare_2019_ICCV} & S     & 89.4  & 43.0  & 82.1  & 30.5  & 21.3  & 30.3  & 34.7  & 24.0  & 85.3  & 39.4  & 78.2  & 63.0  & 22.9  & 84.6  & 36.4  & 43.0  & 5.5   & 34.7  & 33.5  & 46.4  \\
    LSE\cite{Subhani2020LearningFS}   & S     & 90.2  & 40.0  & 83.5  & 31.9  & 26.4  & 32.6  & 38.7  & 37.5  & 81.0  & 34.2  & 84.6  & 61.6  & 33.4  & 82.5  & 32.8  & 45.9  & 6.7   & 29.1  & 30.6  & 47.5  \\
    Ours  & F     & 90.9  & 42.2  & \textbf{83.6} & 33.5  & 22.4  & 30.8  & 34.1  & 24.7  & 83.6  & \textbf{40.8} & 82.9  & 58.9  & 33.2 & 85.1  & \textbf{39.8} & 42.7  & 22.0  & 20.9  & 33.7  & \textbf{47.7} \\
    \midrule
    PatchAlign\cite{Tsai_2019_ICCV} & FS    & 92.3  & 51.9  & 82.1  & 29.2  & 25.1  & 24.5  & 33.8  & 33.0  & 82.4  & 32.8  & 82.2  & 58.6  & 27.2  & 84.3  & 33.4  & 46.3  & 2.2   & 29.5  & 32.3  & 46.5  \\
    BDL\cite{ Li_2019_CVPR}   & IFS   & 91.0  & 44.7  & 84.2  & 34.6  & 27.6  & 30.2  & 36.0  & 36.0  & 85.0  & 43.6  & 83.0  & 58.6  & 31.6  & 83.3  & 35.3  & 49.7  & 3.3   & 28.8  & 35.6  & 48.5  \\
    SIM \cite{SIM_2020_CVPR}  & IFS   & 90.6  & 44.7  & 84.8  & 34.3  & 28.7  & 31.6  & 35.0  & 37.6  & 84.7  & 43.3  & 85.3  & 57.0  & 31.5  & 83.8  & 42.6  & 48.5  & 1.9   & 30.4  & 39.0  & 49.2  \\
    FADA \cite{FADA_ECCV_2020} & FS    & 92.5  & 47.5  & 85.1  & 37.6  & 32.8  & 33.4  & 33.8  & 18.4  & 85.3  & 37.7  & 83.5  & 63.2  & 39.7  & 87.5  & 32.9  & 47.8  & 1.6   & 34.9  & 39.5  & 49.2  \\
    PIT\cite{Lv_PIT_2020_CVPR}   & FS    & 87.5  & 43.4  & 78.8  & 31.2  & 30.2  & 36.3  & 39.9  & 42.0  & 79.2  & 37.1  & 79.3  & 65.4  & 37.5  & 83.2  & 46.0  & 45.6  & 25.7  & 23.5  & 49.9  & 50.6  \\
    FDA \cite{Yang_2020_CVPR}  & IS    & 92.5  & 53.3  & 82.3  & 26.5  & 27.6  & 36.4  & 40.5  & 38.8  & 82.2  & 39.8  & 78.0  & 62.6  & 34.4  & 84.9  & 34.1  & 53.1  & 16.8  & 27.7  & 46.4  & 50.4  \\
    USAMR \cite{2020Unsupervised} & FS    & 90.5  & 35.0  & 84.6  & 34.3  & 24.0  & 36.8  & 44.1  & 42.7  & 84.5  & 33.6  & 82.5  & 63.1  & 34.4  & 85.8  & 32.9  & 38.2  & 2.0   & 27.1  & 41.8  & 48.3  \\
    Ours-ST & FS    & 89.0  & 48.2  & \textbf{85.4} & \textbf{40.6} & 27.0  & 33.4  & 43.8 & 33.0  & \textbf{85.4} & 40.6  & 83.7  & \textbf{63.6} & \textbf{35.3} & 86.9  & 34.3  & 49.9  & \textbf{33.5} & \textbf{35.5} & 33.7  & \textbf{51.7} \\
    \bottomrule
    \end{tabular}}%
  \label{table1}%
\end{table*}%

\begin{table*}[htbp]
  \centering
  \caption{Experimental results for SYNTHIA $\rightarrow$ Cityscapes.
  Consistent with the symbol definition in Table \ref{table1}.
  The mIoU with respect to 13 classes (excluding the “wall”, “fence”, and “pole”) and 16 classes are reported.
  }
  \resizebox{\textwidth}{!}{
    \begin{tabular}{c|c|cccccccccccccccc|c|r} 
    \toprule
    \multicolumn{20}{c}{SYNTHIA-Cityscapes} \\
    \midrule
          & M     & Road  & SW    & Build & Wall  & Fence & Pole  & TL    & TS    & Veg   & Sky   & PR    & Rider & Car   & Bus   & Motor & Bike  & mIoU  & mIoU* \\
    \midrule
    Source Only &  {--}     & 60.5  & 22.1  & 75.2  & 9.3   & 0.2   & 23.7  & 4.7   & 9.0   & 73.0  & 64.1  & 47.3  & 16.1  & 76.1  & 23.8  & 10.9  & 14.3  & 33.1  & 38.2  \\
    \midrule
    SIBAN\cite{Luo_2019_ICCV} & F     & 82.5  & 24.0  & 79.4  & {-}  & {-}  & {-}  & 16.5  & 12.7  & 79.2  & 82.8  & 58.3  & 18.0  & 79.3  & 25.3  & 17.6  & 25.9  & {-}     & 46.3  \\
    CLAN\cite{Luo2019TakingAC}  & F     & 81.3  & 37.0  & 80.1  &{-}  &  {-} & {-}  & 16.1  & 13.7  & 78.2  & 81.5  & 53.4  & 21.2  & 73.0  & 32.9  & 22.6  & 30.7  & {-}  & 47.8  \\
    SSF-DAN\cite{Du_SSFDA_2019_ICCV} & F     & 84.6  & 41.7  & 80.8  & {-}  &  {-} & {-}   & 11.5  & 14.7  & 80.8  & 85.3  & 57.5  & 21.6  & 82.0  & 36.0  & 19.3  & 34.5  & {-}  & 50.0  \\
    FADA \cite{FADA_ECCV_2020}  & F     & 81.5  & 35.1  & 80.6  & 11.5  & 0.3   & 27.0  & 13.5  & 19.5  & 80.9  & 82.0  & 49.4  & 18.6  & 80.2  & 35.3  & 14.2  & 32.0  & 41.4  & 47.8  \\
    CRST\cite{Zou2019ConfidenceRS}  & S     & 67.7  & 32.2  & 73.9  & 10.7  & 1.6   & 37.4  & 22.2  & 31.2  & 80.8  & 80.5  & 60.8  & 29.1  & 82.8  & 25.0  & 19.4  & 45.3  & 43.8  & 50.1  \\
    MaxSquare\cite{ MaxSquare_2019_ICCV} & S     & 82.9  & 40.7  & 80.3  & 10.2  & 0.8   & 25.8  & 12.8  & 18.2  & 82.5  & 82.2  & 53.1  & 18.0  & 79.0  & 31.4  & 10.4  & 35.6  & 41.4  & 48.2  \\
    LSE\cite{Subhani2020LearningFS}   & S     & 82.9  & 43.1  & 78.1  & 9.3   & 0.6   & 28.2  & 9.1   & 14.4  & 77.0  & 83.5  & 58.1  & 25.9  & 71.9  & 38.0  & 29.4  & 31.2  & 42.6  & 49.4  \\
    Ours  & F     & 84.1  & 34.1  & 79.9  & \textbf{11.9} & 0.9   & 28.8  & 19.3  & 24.6  & 81.5  & 80.6  & 53.2  & 23.8  & 78.6  & 32.5  & 17.3  & 44.1  & 43.5  & \textbf{50.3} \\
    \midrule
    PatchAlign\cite{Tsai_2019_ICCV} & FS    & 82.4  & 38.0  & 78.6  & 8.7   & 0.6   & 26.0  & 3.9   & 11.1  & 75.5  & 84.6  & 53.5  & 21.6  & 71.4  & 32.6  & 19.3  & 31.7  & 40.0  & 46.5 \\
    BDL\cite{ Li_2019_CVPR}   & IFS   & 86.0  & 46.7  & 80.3  & {-}  & {-}  & {-}   & 14.1  & 11.6  & 79.2  & 81.3  & 54.1  & 27.9  & 73.7  & 42.2  & 25.7  & 45.3  & {-}   & 51.4 \\
    SIM\cite{SIM_2020_CVPR}   & IFS   & 83.0  & 44.0  & 80.3  & {-} & {-}  & {-}  & 17.1  & 15.8  & 80.5  & 81.8  & 59.9  & 33.1  & 70.2  & 37.3  & 28.5  & 45.8  & {-}  & 52.1 \\
    FADA \cite{FADA_ECCV_2020}  & FS    & 84.5  & 40.1  & 83.1  & 4.8   & 0.0   & 34.3  & 20.1  & 27.2  & 84.8  & 84.0  & 53.5  & 22.6  & 85.4  & 43.7  & 26.8  & 27.8  & 45.2  & 52.5 \\
    PIT\cite{Lv_PIT_2020_CVPR}   & FS    & 83.1  & 27.6  & 81.5  & 8.9   & 0.3   & 21.8  & 26.4  & 33.8  & 76.4  & 78.8  & 64.2  & 27.6  & 79.6  & 31.2  & 31.0  & 31.3  & 44.0  & 51.8 \\
    FDA  \cite{Yang_2020_CVPR}  & IS    & 79.3  & 35.0  & 73.2  & {-}  &{-}  & {-}   & 19.9  & 24.0  & 61.7  & 82.6  & 61.4  & 31.1  & 83.9  & 40.8  & 38.4  & 51.1  & {-}  & 52.5 \\
    USAMR\cite{2020Unsupervised}& FS    & 83.1  & 38.2  & 81.7  & 9.3   & 1.0   & 35.1  & 30.3  & 19.9  & 82.0  & 80.1  & 62.8  & 21.1  & 84.4  & 37.8  & 24.5  & 53.3  & 46.5  & 53.8 \\
    Ours-ST & FS    & \textbf{88.6} & 44.2  & \textbf{83.1} & 2.0   & 0.0   & \textbf{35.4} & 27.7  & 26.7  & \textbf{86.1} & \textbf{86.1} & 61.4  & 28.2  & \textbf{85.6} & \textbf{47.1} & 26.3  & 53.3  & \textbf{48.9} & \textbf{57.3} \\
    \bottomrule
    \end{tabular}}%
  \label{table2}%
\end{table*}%

\begin{figure*}[htp]
    \begin{center}
    \centering 
    \includegraphics[width=1\textwidth, height=0.3\textwidth]{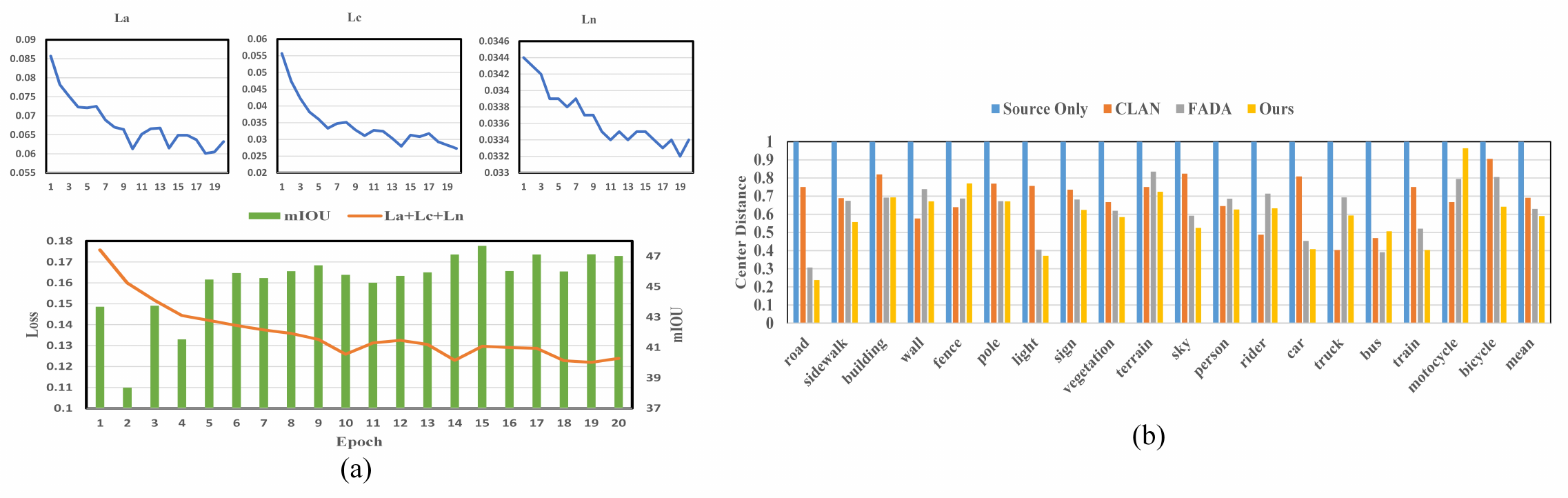} 
    \end{center}
    \caption{(a) Losses and mIOU variation as training goes on. (b) Quantitative analysis of the feature joint distributions. We show the distance of the feature cluster center between the source domain and the target domain for each class.} 
    \label{fig3}
\end{figure*}  

\section{Experiments}
\subsection{Datasets}
We conduct extensive experiments on simulation-to-real unsupervised domain adaptation for the semantic segmentation task, $ e.g. $ SYNTHIA \cite{syhth_dataset} $\rightarrow$ Cityscapes  \cite{Cordts_2016_CVPR} 
and GTA5 \cite{gta5_dataset} $\rightarrow$ Cityscapes.
Cityscapes is a popular urban scene dataset including 2,975 images for training and 500 images for validation.
The resolution of the image is 2048×1024, and pixel-level labels are provided for 19 semantic categories. 
We follow the traditional UDA settings, which uses training set of Cityscapes as the target domain training data and validation set as the testing data. 
GTA5 is a large virtual urban scene datasets rendered by the gaming engine. 
It contains 24999 images, which has the same semantic class as the real scene Cityscapes datasets. 
All the 19 semantic classes are considered in  GTA5 $\rightarrow$ Cityscapes. 
SYNTHIA is another dataset rendered from a virtual city.
A subset of it called SYNTHIA-RAND-CITYSCAPES \cite{syhth_dataset} is utilized, which provides 9,400 images and is paired with Cityscapes. 
And 16 common categories are considered in  SYNTHIA $\rightarrow$ Cityscapes. 

\subsection{Implementation Details}
The network is implemented by PyTorch toolbox. 
For fair comparison with existing works, we adopt the common Deeplab-v2 \cite{deeplab_v2} as our baseline model, ResNet-101 \cite{He_2016_CVPR} as the feature extractor and the $Aspp$ \cite{deeplab_v2} module as the classifier with the sampling rates of $\{6, 12, 18, 24 \}$ . 
The network is pretrained on ImageNet \cite{5206848}. 
The optimizer is Stochastic Gradient Descent (SGD) with momentum of $0.9$ and weight decay of $ 10^{-4} $.
The initial learning rate is set to $2.5 \times 10^{-4}$, and then is  reduced following a poly policy with power of $0.9$.
During data pre-processing, we resize images to the same width (1024) while preserving aspect ratios, then randomly flip them as input.
During training, we warm up the network with 20k iterations on the source domain data, and then use our framework to train another 20k iterations.
The batch size is set as 8. The final $\lambda_{c},  \lambda_{a}$, and $\lambda_{n}$ values are set to 0.001, 0.0015, 0.002 respectively.

After training with our framework, we verify its orthogonality with the self-training strategy \cite{2017A, Zou2019ConfidenceRS}.
Without bells and whistles, we use the segmentation results of the target domain as pseudo-label for self-training, which can further improve performance. 
The detailed results are shown in Table \ref{table1} and \ref{table2}.

\subsection{Comparative Studies}
\textbf{GTA5 $ \rightarrow $ Cityscapes} 
We report the performance from GTA5 to Cityscapes in comparison with existing UDA segmentation methods  \cite{Luo_2019_ICCV, Luo2019TakingAC, Du_SSFDA_2019_ICCV, FADA_ECCV_2020, Zou2019ConfidenceRS, MaxSquare_2019_ICCV, Subhani2020LearningFS, Tsai_2019_ICCV, Li_2019_CVPR, SIM_2020_CVPR, Lv_PIT_2020_CVPR, Yang_2020_CVPR, 2020Unsupervised} in Table \ref{table1}.
Methods using one and multiple adaptation strategies are compared separately. 
Compared with the method that only uses the feature alignment strategy, our method surpass the FADA \cite{ FADA_ECCV_2020} method by $0.9\%$ and the SSF-DAN \cite{ Du_SSFDA_2019_ICCV} method by $2.7\%$, achieving the new state-of-the-art.
It is worth mentioning that both FADA and SSF-DAN introduce additional discriminator, while our method doesn't.
Besides, our method is better than the self-training method MaxSquare\cite{ MaxSquare_2019_ICCV}, CRST\cite{ Zou2019ConfidenceRS}, and LSE \cite{Subhani2020LearningFS}.
In addition, our method and self-training strategy are also complementary.
Combined with the self-training strategy, 
our method further improves $4\%$ and achieves $51.7\%$, 
which tops the score in the existing multi-strategy UDA segmentation methods
\cite{ Tsai_2019_ICCV, Li_2019_CVPR, SIM_2020_CVPR, Lv_PIT_2020_CVPR, Yang_2020_CVPR, 2020Unsupervised}.

\textbf{SYNTHIA  $ \rightarrow $ Cityscapes} 
We also conduct experiments by using SYNTHIA as the source domain. 
The result is reported in Table \ref{table2}.
Compared with the method that only uses the feature alignment strategy, our method surpass the FADA \cite{ FADA_ECCV_2020} method by $2.5\%$ and the SSF-DAN \cite{ Du_SSFDA_2019_ICCV} method by $0.3\%$, achieving the new state-of-the-art.
Compared the self-training methods, our method surpasses the CRST\cite{ Zou2019ConfidenceRS} methods on 13 categories, and is comparable to it on 16 categories. 
The combination with the self-training strategy shows a greater improvement on this task, with an increase of $5.4\%$ on 13 categories and an increase of $7\%$ on 16 categories.
And our method exceeds existing multi-strategy UDA segmentation methods by a large margin in this task. 

\subsection{Training stability}
In this part, two questions are discussed, the convergence of joint training with proposed losses, and the effectiveness of proposed losses adapting to the target domain.
We draw loss curves by epochs and corresponding mIOU column of the validation set in GTA5$ \rightarrow $ Cityscapes task, as shown in Fig.\ref{fig3} (a). 
It can be seen that the $L_a$, $L_c$ and $L_n$ curves drop rapidly during early training, and tend to be stable after 11 epochs. 
The $L_a+L_c+L_n$ loss curve shows a downward trend overall,
which indicates that jointly training three losses can converge stably.
From the total loss curve and the validation mIOU column, 
we can see that the validation mIOU gradually increases as the total loss curve drops. 
This indicates that our proposed method is practicable and adaptive to target domain.

\subsection{Feature distribution}
To further verify the effectiveness of feature alignment, we designed an experiment to show the  alignment degree of each class.
Following \cite{Luo2019TakingAC}, we randomly selected 2k images of the source domain and the target domain, and calculated the cluster center distance(CCD) between domains.
The CCD of the source only model, CLAN \cite{Luo2019TakingAC}, FADA\cite{FADA_ECCV_2020} and our method were calculated for comparison, as shown in Fig.\ref{fig3} (b). 
The CCD of each method is normalized by dividing that of the source only model.
Our method can get a smaller mean CCD value, indicating that the target domain is aligned better than existing state-of-the-art methods. 

\setlength{\unitlength}{1mm}
\begin{figure*}[htp]
    \begin{center}
    \centering 
    \includegraphics[width=1\textwidth, height=0.36\textwidth]{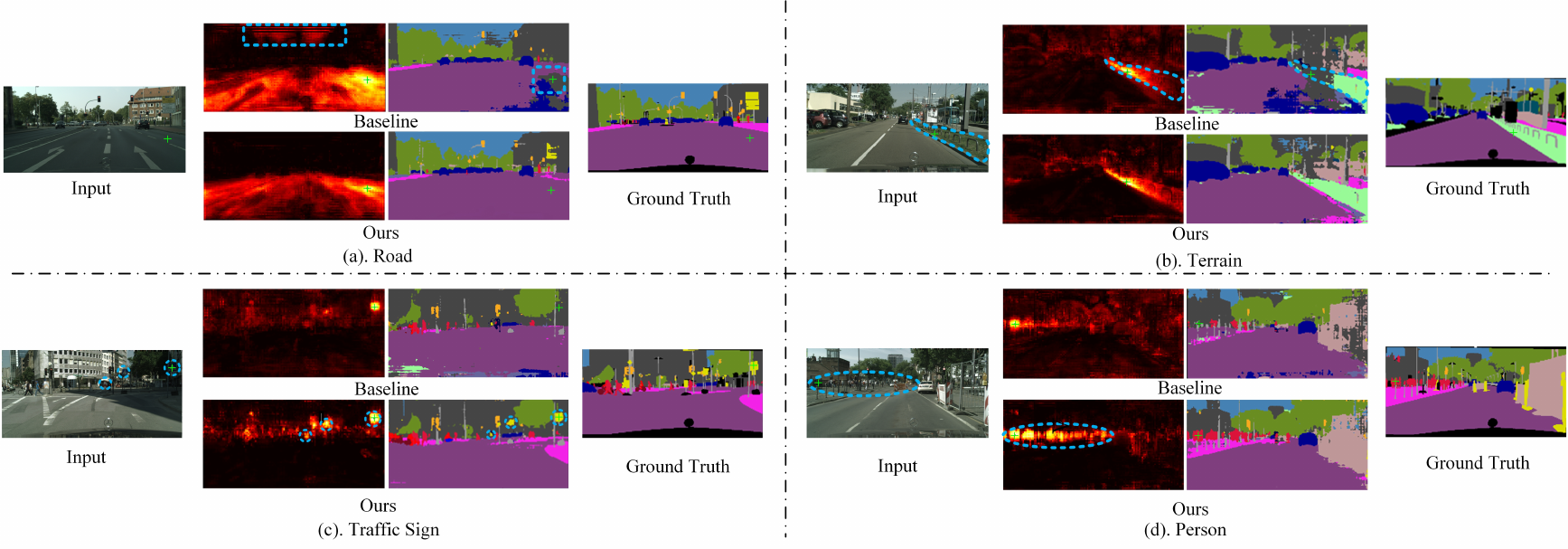} 
    \end{center}
    \caption{Visualization of baseline and our method.  
    For each sub-image, from left to right are the \textbf{input image}, \textbf{feature affinity map}, \textbf{segmentation map} and \textbf{ground truth}.
    The pixel location from sampled class is marked as a green cross in the input picture.
    Feature affinity map is obtained by calculating the feature similarity between this location and the others.
    Brighter the pixel is, more similar the two features are.
    } 
    \label{fig4}
\end{figure*}  

\begin{figure*}[htp]
    \begin{center}
    \centering 
    \includegraphics[width=0.9\textwidth, height=0.18\textwidth]{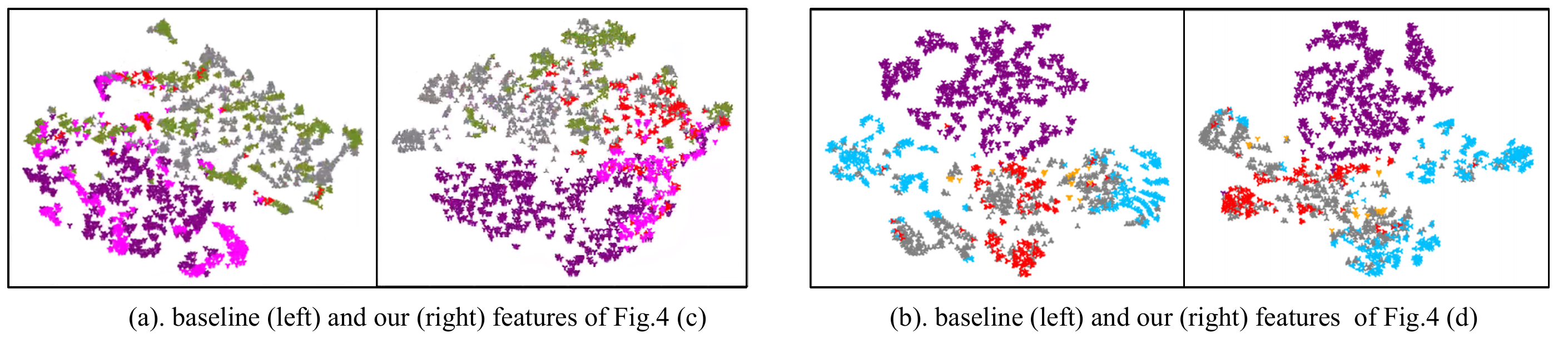} 
    \end{center}
    \caption{Visualize features in Fig.\ref{fig4} (c) and (d) for more class using t-SNE. (For a clear display, we show five classes and the color represents the class in ground truth.) It can be seen that our method can cluster features better than baseline. } 
    \label{fig5}
\end{figure*}

\subsection{Ablation Studies}
In order to evaluate each module, we conduct ablation experiments, as shown in Table \ref{table3}. 
Firstly, $L_a$ is added to $L_{seg}$ to align domains and the results get a primary promotion. 
Then $L_a + L_{seg}$ is regarded as the baseline to evaluate $L_c$ and $L_n$.
Secondly, $L_c$ is added to the baseline and get $3.0\%$ and $1.5\%$ higher for two tasks respectively, which shows that the $L_c$ is effective. 
Thirdly, $L_n$ is added to the baseline and get $3.4\%$ and $2.6\%$ higher for two tasks respectively, which shows that the normalized cut loss $L_n$ is  also effective. 
Finally, the result of $L_{seg} + L_a + L_c + L_n$ shows a huge improvement compared with $L_{seg} + L_a +L_c$ and $L_{seg} + L_a +L_c$, which further illustrates each module is helpful to improve the performance of the segmenter.
In summary, our final performance is $12.1\%$ and $10.7\%$ higher than the source only method (only with $L_{seg}$) on GTA5$ \rightarrow $ Cityscapes and SYNTHIA  $ \rightarrow $ Cityscapes, respectively.

To further verify whether our method can maintain the association among target domain pixels and can adjust the classifier adaptively,  
we visualize the feature affinity graph and segmentation results of GTA5 validation in GTA5 $ \rightarrow $ Cityscapes task, as shown in Fig.\ref{fig4}.
For the former, 
Fig.\ref{fig4} (a) shows that our method can correctly relate to the features with the same semantic of the `road'. 
Fig.\ref{fig4} (c) and (d) show that our method is helpful for enhancing the association among pixels with the same semantic, e.g. `sign' and `person'.
For the latter,
in Fig. \ref{fig4} (a), (b), and (d), although the baseline can maintain the relationship between pixels, the classifier leads to  misclassification.
This means the classifier of baseline doesn't adapt to the target domain.
Compared with baseline, our adaptive classifier assigns the same class label to pixels with high-affinity, and gets a better segmentation results.
As a summary, in our method, the association among pixels with the same semantic can be maintained better and the classifier can be more adaptive. 

Moreover, t-SNE technology \cite{JMLR:v9:vandermaaten08a} is used to visualize the feature distribution of Fig.\ref{fig4}.(c) and (d), as shown in Fig.\ref{fig5}.  It can be seen that the t-SNE results support above discussions. Our method clusters intra-class features and separates inter-class ones better, and those features with high affinity would have the same predicted label.

\begin{table}[htbp]
  \centering
  \caption{Ablation study results of our method.}
    \begin{tabular}{cccc|c|c}
    \toprule
    $L_{seg}$  & $L_{a}$    & $L_{c}$    & $L_{n}$    & GTA5  & SYNTHIA \\
    \midrule
    $\checkmark$     &       &       &       & 35.1  & 33.1  \\
    $\checkmark$     & $\checkmark$     &       &       & 41.7  & 38.8  \\
    $\checkmark$     & $\checkmark$     & $\checkmark$     &       & 44.7  & 40.3  \\
    $\checkmark$     &$\checkmark$       &       & $\checkmark$     & 45.1  & 41.4  \\
    $\checkmark$    & $\checkmark$     & $\checkmark$     & $\checkmark$     & 47.7  & 43.8  \\
    \bottomrule
    \end{tabular}%
  \label{table3}%
\end{table}%

\begin{table}[htbp]
  \centering
  \caption{Influence of $\lambda_c$ given the value of $\lambda_n$ to be stored as 0.002}
    \begin{tabular}{c|ccccc}
    \toprule
    $\lambda_c$ & 0.001 & 0.0015 & 0.002 & 0.003 & 0.004 \\
    \midrule
    mIoU  & 47.3  & 47.7  & 47.2  & 46.8  & 45.4 \\
    \bottomrule
    \end{tabular}%
  \label{table4}%
\end{table}%

\begin{table}[htbp]
  \centering
  \caption{Influence of $\lambda_n$ given the value of $\lambda_c$ to be stored as 0.0015}
    \begin{tabular}{c|ccccc}
    \toprule
    $\lambda_n$ & 0.0005 & 0.001 & 0.002 & 0.003 & 0.004 \\
    \midrule
    mIoU  & 46.9  & 47.3  & 47.7  & 47.4  & 47.1 \\
    \bottomrule
    \end{tabular}%
  \label{table5}%
\end{table}%

\subsection{Parameter Studies}
In this part, we will study the parameters' effect on sensitivity of the algorithm.
The weight $\lambda_a$ of the alignment loss $L_a$ is fixed to 0.001 and the sensitivity of $\lambda_c$ and $\lambda_n$ will be studied.
As shown in Table \ref{table4},
$\lambda_{c}$ with small value is relatively stable, while with large value will cause a drawback. 
We suppose that focusing on clustering very much does not guarantee separability between classes. Because lack of labels for unsupervised clustering, features aren't provided with reasonable semantic guidance from source domain. 
When it comes to $ \lambda_n $ , as shown in Table 5, our model is robust within a certain $\lambda_{n}$ range, which indicates that $L_{n}$ converges stably in a well-clustered feature distribution.

\section{Conclusion}
In this paper, to make the target domain data separable and the classifier adaptive, 
we propose a novel method based on domain closeness assumption for UDA semantic segmentation.
Our method clusters the target domain pixels with the same semantic when alignment and adjusts the classifier to make the decision boundary away from high-density regions.
Our experiments show that each part of our design contributes to the performance gain, which illustrated that our method achieves new state-of-the-arts.

{\small
\bibliographystyle{ieee_fullname}
\bibliography{egbib}
}

\end{document}